\documentclass{article}
\usepackage{natbib}
\usepackage{fullpage}

\usepackage{tikz}
\usepackage[utf8]{inputenc} 
\usepackage{hyperref}       
\usepackage{url}            
\usepackage{booktabs}       
\usepackage{amsfonts}       
\usepackage{amsmath}
\usepackage{nicefrac}       
\usepackage{microtype}      
\usepackage{xcolor}         
\usepackage{graphicx}

\title{SOAK: Same/Other/All K-fold cross-validation for estimating similarity of patterns in data subsets}

\author{%
  Toby Dylan Hocking --- \texttt{toby.dylan.hocking@usherbrooke.ca}\\
  Gabrielle Thibault --- \texttt{gabrielle.thibault.4@ulaval.ca}\\
  Cameron Scott Bodine --- \texttt{CameronBodine@nau.edu}\\
  Paul Nelson Arellano --- \texttt{Paul.Arellano@nau.edu}\\
  Alexander F Shenkin --- \texttt{Alexander.Shenkin@nau.edu}\\
  Olivia Jasmine Lindly --- \texttt{Olivia.Lindly@nau.edu}
}

\begin{document}

\maketitle

\begin{abstract}
In many real-world applications of machine learning, we are interested to know if it is possible to train on the data that we have gathered so far, and obtain accurate predictions on a new test data subset that is qualitatively different in some respect (time period, geographic region, etc). 
Another question is whether data subsets are similar enough so that it is beneficial to combine subsets during model training.
We propose SOAK, Same/Other/All K-fold cross-validation, a new method which can be used to answer both questions.
SOAK systematically compares models which are trained on different subsets of data, and then used for prediction on a fixed test subset, to estimate the similarity of learnable/predictable patterns in data subsets.
We show results of using SOAK on six new real data sets (with geographic/temporal subsets, to check if predictions are accurate on new subsets), 3 image pair data sets (subsets are different image types, to check that we get smaller prediction error on similar images), and 11 benchmark data sets with predefined train/test splits (to check similarity of predefined splits).
\end{abstract}

\section{Introduction}
\label{sec:intro}
A fundamental assumption in supervised learning is similarity between the train data (input to the learning algorithm) and test data (used to evaluate prediction accuracy of the learned model).
This assumption is known as ``independent and identically distributed'' (i.i.d.) in statistics \citep{hastie2009elements}. 
Although special modifications to supervised learning algorithms can guarantee accurate predictions in other scenarios such as covariate shift \citep{sugiyama2012machine}, this paper focuses on standard supervised learning algorithms, designed for i.i.d. data. 
Real-world applications of such supervised learning algorithms often involve training/predicting on data subsets which are qualitatively different in some respect (time period, geographic region, data source, etc.), so the main contribution of this paper is a new method that allows us to quantify if these data subsets are similar enough for accurate learning/prediction.

\begin{figure}
    \centering
    \includegraphics[width=0.7\textwidth]{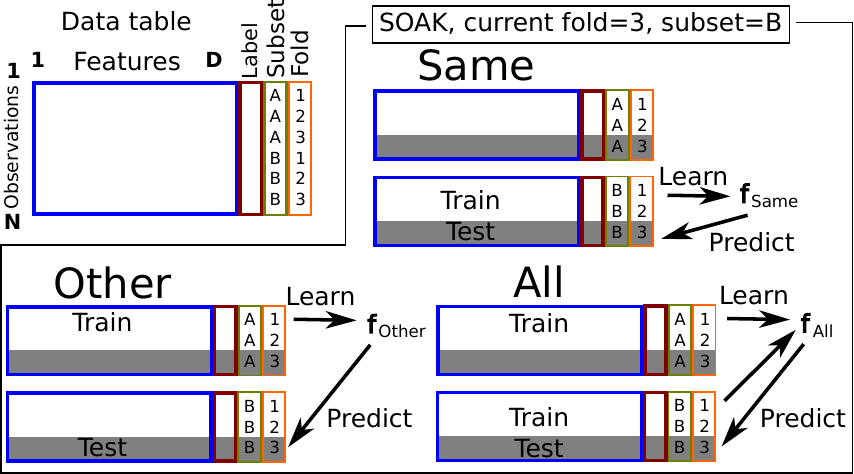}
    \caption{SOAK (Same/Other/All K-fold CV) requires adding subset/fold columns to the data (upper left). 
    For one iteration of SOAK train/test splits (black box, lower right), current test subset=B, so
    Same=B/Other=A/All=A+B are the values of subset which are used to define the train set, in combination with the current fold=3, so test sets shown have subset=B and fold=3, Same train set has subset=B and fold$\in\{1,2\}$, etc.}
    \label{fig:drawing-cv-same-other}
\end{figure}

\paragraph{Motivation for comparing models trained on Same/Other/All subset(s) as test subset.}
For example, in image segmentation, we would like to train on several labeled regions/subsets, and predict on a new region/subset.
In this paper, we use the name ``Other'' for the model trained on several labeled subsets, because it is trained on other subsets of data (relative to the new/test region, see Figure~\ref{fig:drawing-cv-same-other}). 
To determine if those ``Other'' train data subsets are similar/large enough to accurately predict on the new subset, we would like to compare those predictions to a baseline ``Same'' model trained using labeled data from the new subset.
Ideally, these ``Other'' model predictions on a new subset would be just as accurate as ``Same'' model predictions (if learnable/predictable patterns are similar in data subsets), or even more accurate (if the ``Other'' model has access to more training data than the ``Same'' model).

Another real-world example is in learning factors for predicting childhood autism, for which we have two years/subsets of data (2019 and 2020).
We would like to know if it is beneficial to train using ``All'' data subsets (Figure~\ref{fig:drawing-cv-same-other}), because if the subsets are similar enough, then we expect that the larger data set allows the learning algorithm to detect more subtle factors associated with autism (that were not possible to detect by training on any year/subset alone). 
Conversely, if subsets are very different, then we expect that training on ``All'' subsets is detrimental to prediction accuracy (relative to training using the same subset as the test subset). 
Ideally, these ``All'' model predictions on any given subset would be more accurate than ``Same'' model predictions (if patterns in subsets are similar, and each subset is too small when used by itself for training). 

\paragraph{Contributions and novelty.}
In this paper, we propose a new algorithm, Same/Other/All K-fold cross-validation (SOAK), which can be used to compare prediction accuracy of models trained on different data subsets. 
SOAK is a generalization of standard K-fold cross-validation, which corresponds to the special case of SOAK with only one subset, so only the ``Same'' subset as test is used for training.
Although single train/test splits are commonly used for quantifying prediction error on a new subset of data (the ``Other'' model), this method is limited because it only yields a single measurement of prediction error.
Our proposed SOAK algorithm is novel because it combines the idea of K-fold cross-validation, with the idea of training/predicting on qualitatively different data subsets.
There was no existing free software implementation of the proposed SOAK algorithm, so we provide one in the mlr3 framework: \url{https://github.com/tdhock/mlr3resampling}.
After reviewing related work in Section~\ref{sec:related-work}, 
we describe the SOAK algorithm in Section~\ref{sec:methods}, and details of 20 data sets (Table~\ref{tab:meta}).
In Section~\ref{sec:results}, we show results of using SOAK to estimate similarity/differences between subsets in 6 new real data sets (subsets are geographic/temporal), 11 benchmark data sets with 
predefined train and test subsets (which are treated as two SOAK subsets), and 3 image data sets (each has two subsets, MNIST and either EMNIST or FashionMNIST).

\section{Related work}
\label{sec:related-work}
\paragraph{Cross-validation} is a standard method in machine learning, that is widely used, and discussed in several textbooks \citep{bishop2006pattern,hastie2009elements}.
For clarity in this paper, we use the terms from \citet{goodfellow2016deep}: the full data set is split into train/test sets (for evaluation), and then the train set is further split into subtrain/validation sets (for hyper-parameter learning).
K-fold cross-validation can be used for either type of split, and involves partitioning the data into $K$ disjoint test (or validation) sets; for each, the train (or subtrain) set is defined as all the other data.
The idea of averaging several empirical estimators of the risk has been attributed to \citet{geisser1975predictive}.
A primary use of cross-validation is model selection (splitting the train set into subtrain/validation sets), to avoid overfitting during a learning algorithm \citep{Arl_Cel:2010:surveyCV, stephenson2021can}
In contrast, our proposed method is primarily useful for splitting the full data set into train/test sets \citep{ghosh2020approximate}, in order to quantify the prediction error/accuracy on new/test data that were never seen during learning.
Standard K-fold cross-validation can be used for that purpose, and yields $K$ measurements of test error/accuracy that can be useful for comparing the prediction accuracy of different algorithms. 
An alternative is to use a single train/test split, with one subset for train, and another for test; while this approach is somewhat common in the machine learning literature, it only yields one test error/accuracy number, so it can be a misleading estimate of prediction error/accuracy, that tends to encourage overfitting \citep{recht2018cifar}.
In contrast, our proposed SOAK method is based on K-fold CV, so yields $K$ test error/accuracy numbers, and can be used with statistical tests of significance.

\begin{table}[t!]
\centering
\caption{\label{tab:meta}
Meta-data, one row per data set that we analyzed using the proposed SOAK algorithm. 
\\
Imb. = Imbalance ratio between largest/smallest class or subset (1=balanced, larger=more imbalance)}
\small
\begin{tabular}{rllrrrrrr}
  \hline
  & Subset& & & & & class & & subset \\
 & Type & Data & rows & features & classes & Imb. & subsets & Imb. \\ 
  \hline
1 & ImagePair & IPair\_E & 140000 & 784 & 10 & 1.1 &  2 & 1.0 \\ 
  2 & ImagePair & IPair\_E\_rot & 140000 & 784 & 10 & 1.1 &  2 & 1.0 \\ 
  3 & ImagePair & IPair\_Fashion & 140000 & 784 & 10 & 1.1 &  2 & 1.0 \\ 
  4 & time/space & CanadaFiresA & 4827 & 46 &  2 & 2.0 &  4 & 7.0 \\ 
  5 & time/space & CanadaFiresD & 1491 & 46 &  2 & 1.5 &  4 & 1.6 \\ 
  6 & time/space & FishSonar\_river & 2815744 & 81 &  2 & 3.2 &  4 & 1.2 \\ 
  7 & time/space & NSCH\_autism & 46010 & 364 &  2 & 31.8 &  2 & 1.5 \\ 
  8 & time/space & aztrees3 & 5956 & 21 &  2 & 7.8 &  3 & 2.0 \\ 
  9 & time/space & aztrees4 & 5956 & 21 &  2 & 7.8 &  4 & 4.9 \\ 
  10 & train/test & CIFAR10 & 60000 & 3072 & 10 & 1.0 &  2 & 5.0 \\ 
  11 & train/test & EMNIST & 70000 & 784 & 10 & 1.0 &  2 & 6.0 \\ 
  12 & train/test & FashionMNIST & 70000 & 784 & 10 & 1.0 &  2 & 6.0 \\ 
  13 & train/test & KMNIST & 70000 & 784 & 10 & 1.0 &  2 & 6.0 \\ 
  14 & train/test & MNIST & 70000 & 784 & 10 & 1.2 &  2 & 6.0 \\ 
  15 & train/test & QMNIST & 120000 & 784 & 10 & 1.2 &  2 & 1.0 \\ 
  16 & train/test & STL10 & 13000 & 27648 & 10 & 1.0 &  2 & 1.6 \\ 
  17 & train/test & spam & 4601 & 57 &  2 & 1.5 &  2 & 2.0 \\ 
  18 & train/test & vowel & 990 & 10 & 11 & 1.0 &  2 & 1.1 \\ 
  19 & train/test & waveform & 800 & 21 &  3 & 1.1 &  2 & 1.7 \\ 
  20 & train/test & zipUSPS & 9298 & 256 & 10 & 2.2 &  2 & 3.6 \\ 
   \hline
\end{tabular}
\end{table}

\paragraph{Distributional Homogeneity.}
The cross-validation method that we propose is related to the statistical concepts of independent and identically distributed (i.i.d.) random variables, and of homogeneity in meta-analyses, meaning that different subsets of the data follow the same distribution.
Homogeneity/i.i.d. is a stronger condition than we are interested in measuring using our proposed cross-validation procedure (it may be beneficial to combine subsets when learning, even though they are heterogeneous).
Classic examples of statistical tests for homogeneity are the Chi-Square test of \citet{Pearson1900} and the Q test of \citet{cochran1954combination}.
In meta-analysis, the goal is to provide a better estimate of a quantity measured in several different studies, and there are several methods available for estimating heterogeneity \citep{veroniki2016methods}.
The ReDistribution test of homogeneity has been used for climate time series data \citep{romanic2015long}, and is related to the study of change-point detection algorithms \citep{truong2020selective}.

\paragraph{Novelty with respect to available software.}
There are many free/open-source implementations of cross-validation, including origami \citep{coyle2018origami}, splitTools \citep{Mayer2023splitTools} and mlr3 \citep{Lang2019mlr3} in R, as well as scikit-learn in python \citep{scikit-learn}.
All of these packages support standard K-fold cross-validation, and some support stratification, and keeping groups of observations together when splitting (a concept/parameter called ``group'').
The proposed SOAK algorithm is based on the concept of data subsets, which was not previously supported in any free software machine learning framework, so we implemented it in the mlr3 framework (\url{https://github.com/tdhock/mlr3resampling}), because it provides flexible support for parallelization over data sets, algorithms, and train/test splits (including parallelization over the proposed subsets).
The ``group'' concept (observations/rows in the same group must be assigned the same fold ID) is present in both mlr3 and in scikit-learn, and the ``group'' concept is different from the SOAK ``subset'' concept (one subset is designated as test, and Same/Other/All subsets are designated as train), but actually the two concepts can be used together. 
For example, in image segmentation, labels are often created by drawing polygons, and in each polygon there are several pixels which are assigned the same label. If each polygon is considered a group, then each pixel in a polygon gets the same group ID. The image(s) may be divided into regions/subsets such as North/South/etc in the aztrees data. So we can use both concepts at the same time: let labeled pixels from polygons/groups in the South region/subset be the test set, and define the train set as labeled pixels from polygons/groups in the Same/Other/All region/subset.




\section{Methods: proposed SOAK algorithm and data sets}
\label{sec:methods}

In this section, we give details of the proposed SOAK algorithm, and then give details about the 20 data sets we analyzed (Table~\ref{tab:meta}), which represent classification problems with 800--2,816,744 rows, 10--27,648 features, 2--11 classes, and 2--4 subsets.
\subsection{Proposed SOAK algorithm}
We propose SOAK (Same/Other/All K-fold CV), a new variant of cross-validation which is useful for determining similarity between learnable/predictable patterns in data subsets.
As in standard K-fold cross-validation for supervised machine learning, we assume there is a data set with $N$ observations/rows.
Additionally, we assume that the rows can be partitioned into a certain number of subsets $S$, and we would like to estimate the similarity of learnable/predictable patterns these subsets.
Each subset has an identifier, which we treat here as an integer from 1 to $S$.
For example in Figure~\ref{fig:drawing-cv-same-other} there are $S=2$ subsets, A and B.
For each observation/row $i$, we assume there is a corresponding subset $s_i\in\{1,\dots,S\}$ and fold ID $k_i\in\{1,\dots,K\}$ (assigned randomly or using strata, such that each label/subset has about the same number of rows with a given fold ID).

The goal of SOAK is to estimate the prediction error/accuracy of a learning algorithm, when attempting to predict on a given subset, and training on that same subset, or on different subsets (others or all).
Therefore, our method has a loop over each subset $\sigma\in\{1,\dots,S\}$ and fold $\kappa\in\{1,\dots,K\}$. 
For each, we define the train/test splits as in Figure~\ref{fig:drawing-cv-same-other}. 
\begin{description}
  \item[Test] set is 
  $
  \{i\in\{1,\dots,N\}\mid k_i= \kappa\text{ and }s_i=\sigma\}$, 
  all rows $i$ in the current fold $\kappa$ and subset $\sigma$.
  \item[Same] train set is 
  $
  \{i\in\{1,\dots,N\}\mid k_i\neq \kappa\text{ and }s_i=\sigma\}$, 
  all rows $i$ not in the current fold $\kappa$, but in the current subset $\sigma$.
  \item[Other] train set is 
  $
  \{i\in\{1,\dots,N\}\mid k_i\neq \kappa\text{ and }s_i\neq\sigma\}$, 
  all rows $i$ which are neither in the current fold $\kappa$, nor in the current subset $\sigma$.
  \item[All] train set is 
  $
  \{i\in\{1,\dots,N\}\mid k_i\neq \kappa\}$, 
  all rows $i$ not in the current fold $\kappa$.
\end{description}
SOAK runs learning algorithm(s) on Same/Other/All train sets, then computes the resulting predictions on the Test set for this subset $\sigma$ and fold $\kappa$.
Note that the subset column is not included as a feature for learning/prediction.
This is repeated for each fold $\kappa$, yielding $K$ measures of test error/accuracy for each train set (Same/Other/All), when predicting on a given subset $\sigma$.
Although other significance tests have been proposed \citep{bayle2020cross}, we propose for simplicity to perform two comparisons using a paired t-test with $K-1$ degrees of freedom:
\begin{description}
    \item[Same vs Other:] if test accuracy for Other is comparable or better than Same, then we can conclude that it is possible to train on other subsets, and accurately predict on this subset.
    \item[Same vs All:] if test accuracy for All is significantly better than Same, then we can conclude that combining subsets when training is beneficial to prediction accuracy in this subset. 
\end{description}
This is repeated for each subset $\sigma$, yielding $S$ t-test results (mean error differences and p-values, for Same/Other and Same/All).

\paragraph{Standard CV is a special case.}
Note that standard K-fold CV is the special case with $S=1$ subset, which means that the Other train set is empty, the All train set is identical to Same, and so for each learning algorithm, we have only $K$ test accuracy/error numbers (one for each fold/split).

\paragraph{Computational complexity.}
For each of $S$ test subsets and $K$ folds, we need to consider training on Same/Other/All subsets, so the number of train/test splits considered by SOAK is $3SK=O(SK)$, which is the number of times each learning algorithm needs to be run.
Importantly, this is linear in the number of subsets $S$, so it is possible to run SOAK on data with a large number of subsets $S$.

\paragraph{Implementation Details.} 
SOAK can be easily implemented in any programming language, by looping over all subsets $\sigma\in\{1,\dots,S\}$ and fold IDs $\kappa\in\{1,\dots,K\}$.
We implemented the computational experiments in this paper using the mlr3 framework in R, which made it easy to compute results in parallel (over all algorithms, data sets, and train/test splits) using the mlr3batchmark package \citep{mlr3batchmark}.

\subsection{Image Pairs: train on MNIST, predict on EMNIST or FashionMNIST}
The MNIST data set consists of 70,000 images of handwritten digits, and the goal of learning is to accurately classify each image into one of ten classes (0 to 9) \citep{lecun1998gradient}.
EMNIST is another set of 70,000 images, also of handwritten digits, with balanced classes (see Table~\ref{tab:meta}, column class Imb.), and a different pre-processing technique that attempts to scale images to fill the available space \citep{EMNIST}.
FashionMNIST is a set of 70,000 images, with 7,000 examples of each of ten classes of clothing \citep{xiao2017fashion}.
Intuitively, we expect that we should be able to train on MNIST (images of digits), and get reasonable predictions on EMNIST (because images are also digits), but not on FashionMNIST (because images are clothing).
Therefore, we created three data sets (Table~\ref{tab:meta}), each with 14,000 images, by combining MNIST with one of the other variants.
\begin{description}
    \item[IPair\_E:] MNIST combined with EMNIST. 
    Note that the raw EMNIST images not in the same orientation as MNIST (if MNIST is upright, then EMNIST appears to be rotated 90°).
    \item[IPair\_E\_rot:] MNIST combined with rotated EMNIST (all images in upright orientation).
    \item[IPair\_Fashion:] MNIST combined with FashionMNIST.
\end{description}
In each of the three data sets, there are two subsets (one from each source: MNIST and EMNIST or FashionMNIST).

\subsection{Benchmark data with a predefined train/test split}

There are many benchmark data sets in the machine learning literature that include a column to designate a predefined train/test split.
In this paper, we consider several data sets which were downloaded using torchvision \citep{marcel2010torchvision}: CIFAR10 \citep{alex2009learning}, EMNIST \citep{EMNIST}, FashionMNIST \citep{xiao2017fashion}, KMNIST \citep{KMNIST}, MNIST \citep{lecun1998gradient}, QMNIST \citep{qmnist-2019}, STL10 \citep{STL10}; several others that are included as supplementary materials in the textbook of \citet{hastie2009elements} (spam, vowel, waveform), and one data set that was included in both (USPS in torch, and zip in the textbook, which we call zipUSPS in this paper).
Rather than using the predefined train/test split for its intended purpose, we instead use it as a subset ID, so each of these data sets has two subsets (Table~\ref{tab:meta}).
By using SOAK on these data, we seek to answer the question: are the predefined train and test subsets similar enough, so that if train on one subset (either predefined=train or test), we can get predictions on the other subset, that are just as accurate as if we had access to the same subset?

\subsection{Real data from various application domains with spatial/temporal subsets}

\paragraph{CanadaFires} data consist of four satellite images of forest fires, in which we are primarily interested to answer the question: can we train on some images, and accurately predict burned areas in a new image? (subsets are images of different forest fires)
Characterizing the burning pattern using these high-resolution Skysat images is important because it is used by the Québec government to plan salvage operations. 
Skysat images of forest fires that occurred in 2020--2021 were used. In addition, Landsat were obtained using Google Earth Engine  \citep{gorelick2017google}. 
Each row/observation in these data is a labeled pixel, and there are columns/features for Normalized Burn Ratio \citep{french2008using}, delta
Normalized Burn Ratio \citep{parks2014new}, Normalized Difference Vegetation Index \citep{pettorelli2013normalized}, Green Chromatic Coordinates \citep{alberton2023relationship}, etc., for a total of 46 features.  
Labels were created for individual pixels, each at least 100m apart, and more than 6m from the image borders, by manually assigning one of six burn classes, based on a classification that the Québec government uses to characterize fires.
We then transformed the labels to a binary problem, burned (positive class)
versus other (negative class). 
There are two versions of these data: A means All data, and D means Down-sampled to
promote class balance, while retaining representative examples. 
In these data, we are interested to see if it is possible to train on a few images/fires, and accurately predict on a new image/fire (there are four such images/subsets).

\paragraph{aztrees} 
data come from satellite images around Flagstaff, AZ, in which we are primarily interested to answer this question: can we train on some regions, and accurately predict presence of trees in new regions? 
(subsets are different regions in the satellite image)
The goal is to predict presence/absence of trees throughout Arizona, in a project that seeks to determine the extent to which trees are under stress/drought/bark beetle infestation.
Three sets of satellite images were retrieved from Google Earth Engine: Sentinel-1, Sentinel-2, and the NASA Shuttle Radar Topography Mission (SRTM) Global image \citep{NASA_SRTM,gorelick2017google}.
Data were converted so that each row is a pixel, and 20 features/columns were computed, including several spectral bands, Normalized Difference Vegetation Index-NDVI \citep{NDVI}, Normalized Difference Infrared Index-NDII \citep{NDII}, Normalized Difference Water Index-NDWI \citep{NDWI}, three Topographic Position-TPI Indexes \citep{TPI}.
Images were manually labeled using Quantum GIS software \citep{QGIS_software} by drawing polygons that indicate seven Land Use/Land Cover classes (trees, natural grass, planted grass, infrastructure, bare ground, and open water), then labels were converted into a binary problem (tree versus other).
There are two variants of these data, with either 3 (S/NE/NW) or 4 (SE/SW/NE/NW) geographic subsets, and we would like to know if it is possible to train on some subsets, and accurately predict on a new subset.

\paragraph{FishSonar\_river} data come from sonar imagery of river bottoms, in which we are primarily interested to answer this question: can we train on a few rivers, and accurately predict areas suitable for fish spawning, on a new river? The goal is to accurately predict spawning areas of Gulf Sturgeon (\textit{Acispenser oxyrinchus desotoi}), in a wildlife conservation project \citep{U.S.FishandWildlifeService2022}. 
In detail, data come from Humminbird\textsuperscript{\textregistered} fish finder sonar from surveys in the Pearl and Pascagoula watersheds in Mississippi, USA over 2021--2023, then processed with PING-Mapper software \citep{Bodine2022a, Bodine2022b, Bodine2023_ping}.
Sonar data give imagery of the river bottom; the goal is to classify each pixel as either suitable (gravel, cobbles, boulders, and bedrock) or not (silt, mud, and sand), for Gulf Sturgeon spawning \citep{Sulak2016}.
Data were converted so that each row is a pixel, and each column is a mean over windows around that pixel (mean pooling, window size 9).
Multiple class labels were first created using Doodler software \citep{Buscombe2022a}, and then transformed to obtain a binary classification problem: hard bottom (suitable for spawning), versus anything else.
Subsets to train/test on were defined using four different rivers (Bouie, Chickasawhay, Leaf, Pearl), and we are interested to see if models learned on a subset of rivers can be used for accurate prediction on a new river.

\begin{figure}
    \centering
    \includegraphics[width=\textwidth]{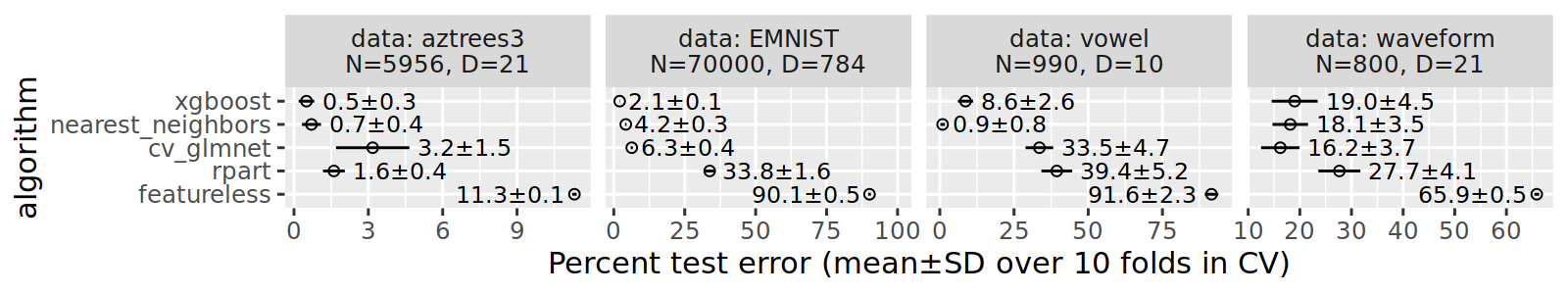}
    
    \includegraphics[width=\textwidth]{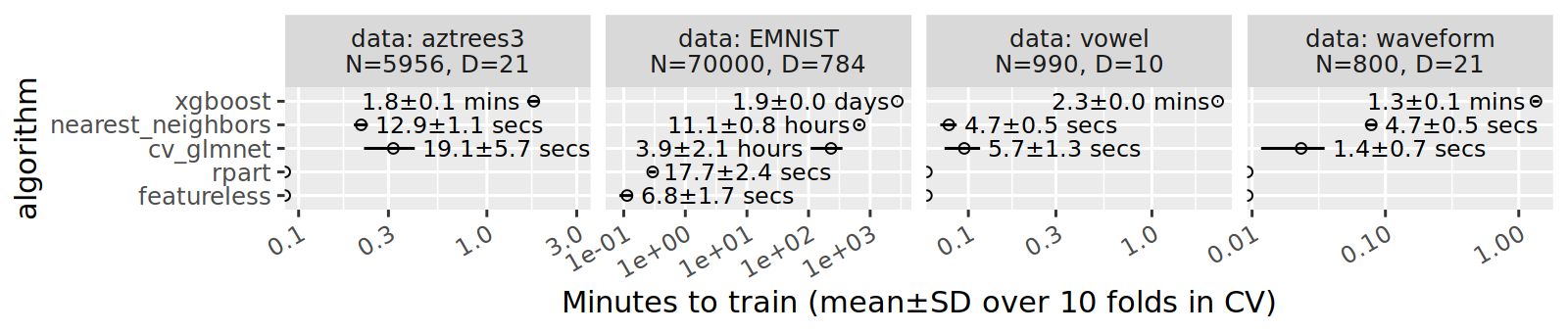}
    \vspace{-0.3in}
    \caption{
    Test error (top) and training time (bottom) of five algorithms on four data sets. 
    }
    \label{fig:five_algos}
\end{figure}

\paragraph{NSCH\_autism data} consist of two subsets/years (2019 and 2020) from the National Survey of Children's Health (NCSH) \citep{ghandour2018design,NSCH2019, NSCH2020}.
The goal of this machine learning analysis is to accurately predict autism diagnosis, using an interpretable model which can identify a subset of questions/responses that are useful for prediction.
Each row represents a child for whom an adult familiar with the child's health answered the survey in a given year, and each column represents a question/response with various categories such as mental health diagnoses, family income, health insurance status, healthcare needed and received, neighborhood characteristics, etc. 
There are two subsets/years in these data, and we would like to know if training with combined years results in a more accurate model. 

\section{Results: estimating differences between subsets in 20 data sets}
\label{sec:results}
In this section, we show how SOAK can be used to quantify similarity/differences between subsets in 20 data sets.
We ran experiments in parallel using a compute cluster: one CPU per data set, algorithm, and train/test split (64GB RAM and 2 days computation time for each).

\subsection{Comparing prediction accuracy of 5 learning algorithms on 20 data sets}
First, we used standard 10-fold cross-validation on 20 classification data sets (Table~\ref{tab:meta}) to compare the prediction accuracy of 5 learning algorithms (with hyper-parameters tuned using internal 10-fold cross-validation): cv\_glmnet (L1-regularized linear model); featureless (baseline always predicting most frequent class); rpart (default decision tree learner with no hyper-parameter tuning); nearest neighbors (tuned 1 to 20 neighbors); xgboost (gradient boosting, learning rate \texttt{eta} tuned over 5 values on log scale between 0.001 and 1, \texttt{nrounds} tuned over 5 values from 1 to 100).
From Figure~\ref{fig:five_algos}, we see that  xgboost was always the slowest, and sometimes most accurate; cv\_glmnet had reasonable/intermediate computation time and prediction accuracy, so we used it for the other experiments with SOAK.

\begin{figure}
    \centering
    \includegraphics[width=0.9\textwidth]{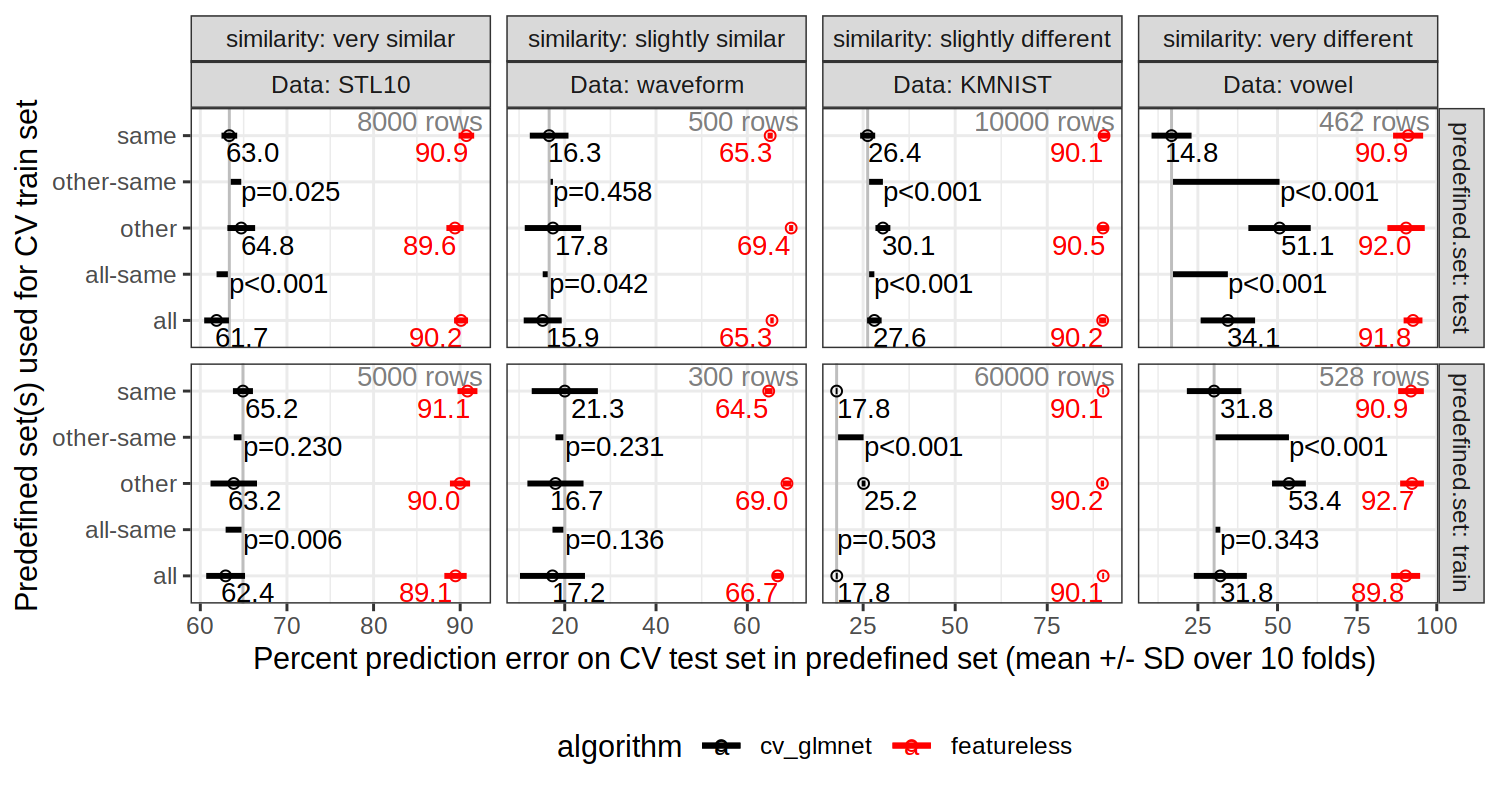}
    \vspace{-0.3in}
    \caption{
    SOAK was used to compute mean/SD of test error over 10 cross-validation folds, and p-values for differences (other-same and all-same), in each of four data sets in which there were two subsets (predefined train/test assignments in the data table).
    For data sets that have similar learnable/predictable patterns (left), training on all subsets has smaller test error than same, and training on other has either smaller or larger test error than same (depending on number of rows in subset).
    For data sets that have different learnable/predictable patterns (right), training on all subsets never has smaller test error than same, and training on other always has larger test error than same.
    }
    \label{fig:four_train_test}
\end{figure}

\subsection{SOAK prediction error comparison plots}
Next, we used SOAK with 10-fold CV on each of the 20 data sets, to compute prediction error on each subset, after training cv\_glmnet on Same/Other/All subset(s).
We expected that some data sets would have very similar subsets (All model better than Same), and others would have very different subsets (Same model better than All).
In particular, since train/test splits are typically assigned randomly (due to the i.i.d. assumption), we expected to observe large similarity in train/test subsets, such as in STL10, waveform, KMNIST, and vowel data sets (Figure~\ref{fig:four_train_test}).
Two of those data sets (STL10 and waveform) exhibited some evidence of similarity between subsets: All test error was smaller than Same test error (STL10 significantly, $p<0.05$ in two-sided paired $t_9$-tests; waveform slightly, $p=0.042$ and 0.136).
Additional evidence of similarity between subsets in STL10 was that Other test error was significantly smaller or larger than Same (depending on sample size, predefined train subset had 8000 rows whereas predefined test subset had only 5000 rows). 
Surprisingly, there were two data sets (KMNIST and vowel) which exhibited differences between learnable/predictable patterns in predefined train/test subsets.
Other test error was significantly larger than Same (mean difference of 5--25\%, $p<0.001$), and All test error was significantly larger than Same for the smaller subset (predefined test subset, $p<0.001$).
Overall, it is clear that by running SOAK then plotting Same/Other/All test error (Figure~\ref{fig:four_train_test}), it is possible to estimate the extent of similarity/differences between learnable/predictable patterns in data subsets.

\begin{figure}[t]
    \centering
    \parbox{0.61\textwidth}{
    \includegraphics[width=0.59\textwidth]
    {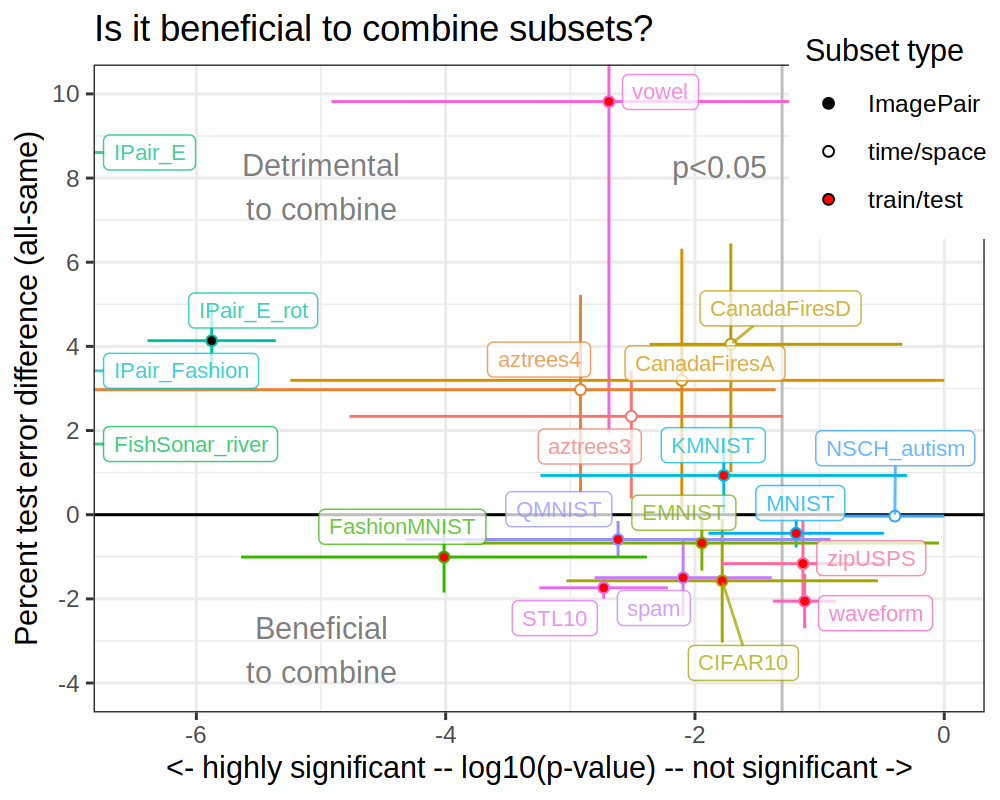}
    }
    \parbox{0.38\textwidth}{
    \scriptsize
\begin{tabular}{lll}
  \hline
Data & ErrorDiff & log10(P) \\ 
  \hline
\tikz\draw[black,fill=black] (0,0) circle (.5ex); IPair\_E & $\parbox{0.5cm}{\rightline{8.2}},\parbox{0.5cm}{\rightline{9.1}}$ & $\parbox{0.5cm}{\rightline{-11.2}},\parbox{0.5cm}{\rightline{-10.2}}$ \\ 
  \tikz\draw[black,fill=black] (0,0) circle (.5ex); IPair\_E\_rot & $\parbox{0.5cm}{\rightline{3.3}},\parbox{0.5cm}{\rightline{5.0}}$ & $\parbox{0.5cm}{\rightline{-6.4}},\parbox{0.5cm}{\rightline{-5.4}}$ \\ 
  \tikz\draw[black,fill=black] (0,0) circle (.5ex); IPair\_Fashion & $\parbox{0.5cm}{\rightline{2.6}},\parbox{0.5cm}{\rightline{4.6}}$ & $\parbox{0.5cm}{\rightline{-12.4}},\parbox{0.5cm}{\rightline{-8.6}}$ \\ 
  \tikz\draw[black,fill=red] (0,0) circle (.5ex); vowel & $\parbox{0.5cm}{\rightline{2.0}},\parbox{0.5cm}{\rightline{17.7}}$ & $\parbox{0.5cm}{\rightline{-4.9}},\parbox{0.5cm}{\rightline{-0.5}}$ \\ 
  \tikz\draw[black,fill=white] (0,0) circle (.5ex); CanadaFiresD & $\parbox{0.5cm}{\rightline{1.0}},\parbox{0.5cm}{\rightline{6.4}}$ & $\parbox{0.5cm}{\rightline{-2.4}},\parbox{0.5cm}{\rightline{-0.3}}$ \\ 
  \tikz\draw[black,fill=white] (0,0) circle (.5ex); aztrees4 & $\parbox{0.5cm}{\rightline{0.5}},\parbox{0.5cm}{\rightline{5.2}}$ & $\parbox{0.5cm}{\rightline{-7.0}},\parbox{0.5cm}{\rightline{-1.4}}$ \\ 
  \tikz\draw[black,fill=white] (0,0) circle (.5ex); FishSonar\_river & $\parbox{0.5cm}{\rightline{0.5}},\parbox{0.5cm}{\rightline{4.4}}$ & $\parbox{0.5cm}{\rightline{-14.4}},\parbox{0.5cm}{\rightline{-8.6}}$ \\ 
  \tikz\draw[black,fill=white] (0,0) circle (.5ex); aztrees3 & $\parbox{0.5cm}{\rightline{0.4}},\parbox{0.5cm}{\rightline{3.4}}$ & $\parbox{0.5cm}{\rightline{-4.8}},\parbox{0.5cm}{\rightline{-1.3}}$ \\ 
  \tikz\draw[black,fill=red] (0,0) circle (.5ex); KMNIST & $\parbox{0.5cm}{\rightline{0.0}},\parbox{0.5cm}{\rightline{1.8}}$ & $\parbox{0.5cm}{\rightline{-3.2}},\parbox{0.5cm}{\rightline{-0.3}}$ \\ 
  \tikz\draw[black,fill=white] (0,0) circle (.5ex); CanadaFiresA & $\parbox{0.5cm}{\rightline{-0.0}},\parbox{0.5cm}{\rightline{6.3}}$ & $\parbox{0.5cm}{\rightline{-5.2}},\parbox{0.5cm}{\rightline{0.0}}$ \\ 
  \hline
  \tikz\draw[black,fill=white] (0,0) circle (.5ex); NSCH\_autism & $\parbox{0.5cm}{\rightline{-0.1}},\parbox{0.5cm}{\rightline{-0.0}}$ & $\parbox{0.5cm}{\rightline{-0.8}},\parbox{0.5cm}{\rightline{-0.0}}$ \\ 
  \tikz\draw[black,fill=red] (0,0) circle (.5ex); MNIST & $\parbox{0.5cm}{\rightline{-0.8}},\parbox{0.5cm}{\rightline{-0.1}}$ & $\parbox{0.5cm}{\rightline{-1.9}},\parbox{0.5cm}{\rightline{-0.5}}$ \\ 
  \tikz\draw[black,fill=red] (0,0) circle (.5ex); QMNIST & $\parbox{0.5cm}{\rightline{-1.0}},\parbox{0.5cm}{\rightline{-0.2}}$ & $\parbox{0.5cm}{\rightline{-4.3}},\parbox{0.5cm}{\rightline{-0.9}}$ \\ 
  \tikz\draw[black,fill=red] (0,0) circle (.5ex); EMNIST & $\parbox{0.5cm}{\rightline{-1.3}},\parbox{0.5cm}{\rightline{-0.0}}$ & $\parbox{0.5cm}{\rightline{-3.8}},\parbox{0.5cm}{\rightline{-0.0}}$ \\ 
  \tikz\draw[black,fill=red] (0,0) circle (.5ex); FashionMNIST & $\parbox{0.5cm}{\rightline{-1.9}},\parbox{0.5cm}{\rightline{-0.2}}$ & $\parbox{0.5cm}{\rightline{-5.6}},\parbox{0.5cm}{\rightline{-2.4}}$ \\ 
  \tikz\draw[black,fill=red] (0,0) circle (.5ex); STL10 & $\parbox{0.5cm}{\rightline{-2.0}},\parbox{0.5cm}{\rightline{-1.5}}$ & $\parbox{0.5cm}{\rightline{-3.2}},\parbox{0.5cm}{\rightline{-2.2}}$ \\ 
  \tikz\draw[black,fill=red] (0,0) circle (.5ex); zipUSPS & $\parbox{0.5cm}{\rightline{-2.2}},\parbox{0.5cm}{\rightline{-0.1}}$ & $\parbox{0.5cm}{\rightline{-1.8}},\parbox{0.5cm}{\rightline{-0.5}}$ \\ 
  \tikz\draw[black,fill=red] (0,0) circle (.5ex); spam & $\parbox{0.5cm}{\rightline{-2.4}},\parbox{0.5cm}{\rightline{-0.6}}$ & $\parbox{0.5cm}{\rightline{-2.8}},\parbox{0.5cm}{\rightline{-1.4}}$ \\ 
  \tikz\draw[black,fill=red] (0,0) circle (.5ex); waveform & $\parbox{0.5cm}{\rightline{-2.7}},\parbox{0.5cm}{\rightline{-1.4}}$ & $\parbox{0.5cm}{\rightline{-1.4}},\parbox{0.5cm}{\rightline{-0.9}}$ \\ 
  \tikz\draw[black,fill=red] (0,0) circle (.5ex); CIFAR10 & $\parbox{0.5cm}{\rightline{-3.0}},\parbox{0.5cm}{\rightline{-0.1}}$ & $\parbox{0.5cm}{\rightline{-3.0}},\parbox{0.5cm}{\rightline{-0.5}}$ \\ 
   \hline
\end{tabular}
}
    \caption{
    SOAK was used to compute mean test error differences (All-Same) and p-values for each test subset, over 10 cross-validation folds.
    Line segments and table show min/max values over 2--4 test subsets in each data set; dot shows mean.
    Horizontal black line separates data sets by the degree of differences in learnable/predictable patterns: top 10 for large differences (min/max ErrorDiff positive or zero: never beneficial to combine subsets when training) and bottom 10 for small differences (min/max ErrorDiff negative or zero: never detrimental to combine subsets).
    }
    \label{fig:map_all}
\end{figure}
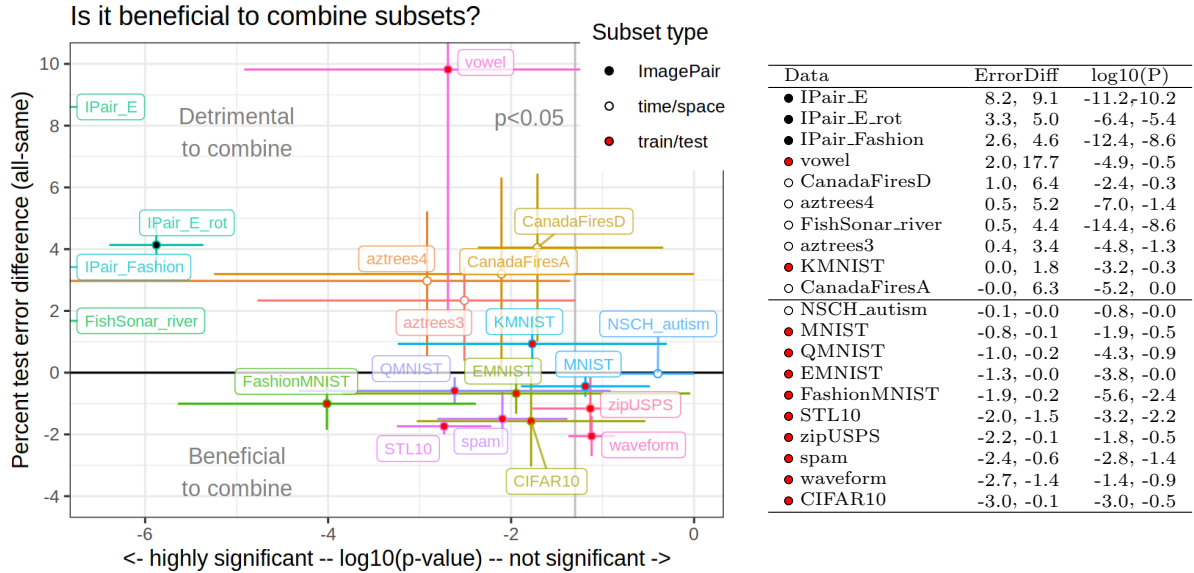

\subsection{SOAK scatterplots of test error differences and p-values}
After running SOAK on each of the 20 data sets, we plotted mean test error differences (All-Same, Figure~\ref{fig:map_all}; Other-Same, Figure~\ref{fig:map_other}), and p-values (two-sided paired $t_9$-test because we used 10-fold CV).
Each plot/table shows a line/row representing the min/max test error difference and p-value, over the 2--4 subsets in each data set.
When analyzing All-Same test error differences (Figure~\ref{fig:map_all}), it is clear that the data sets can be divided into two categories: 10 data sets have similar subsets (negative test error differences, All-Same) and the other 10 have different subsets (zero or positive test error differences). 
Similarly when analyzing Other-Same test error differences (Figure~\ref{fig:map_other}),
we obtain the same categorization of 10 data sets with similar subsets (min Other-Same test error difference negative, max positive), and 10 data sets with different subsets (min/max test error difference both positive).
As expected, real data sets with space/time subsets, and ImagePair data, tended to have different subsets, whereas data with predefined train/test subsets were mostly similar (exceptions were NSCH\_autism real data with slight similarity of subsets/years, and vowel/KMNIST with different predefined train/test subsets).
Overall, these SOAK scatterplots/tables allow categorization of data sets as having subsets with either similar or different learnable/predictable patterns.

\section{Discussion and Conclusions}
\label{sec:discussion-conclusions}

\subsection{Image pairs: MNIST data combined with EMNIST/FashionMNIST}

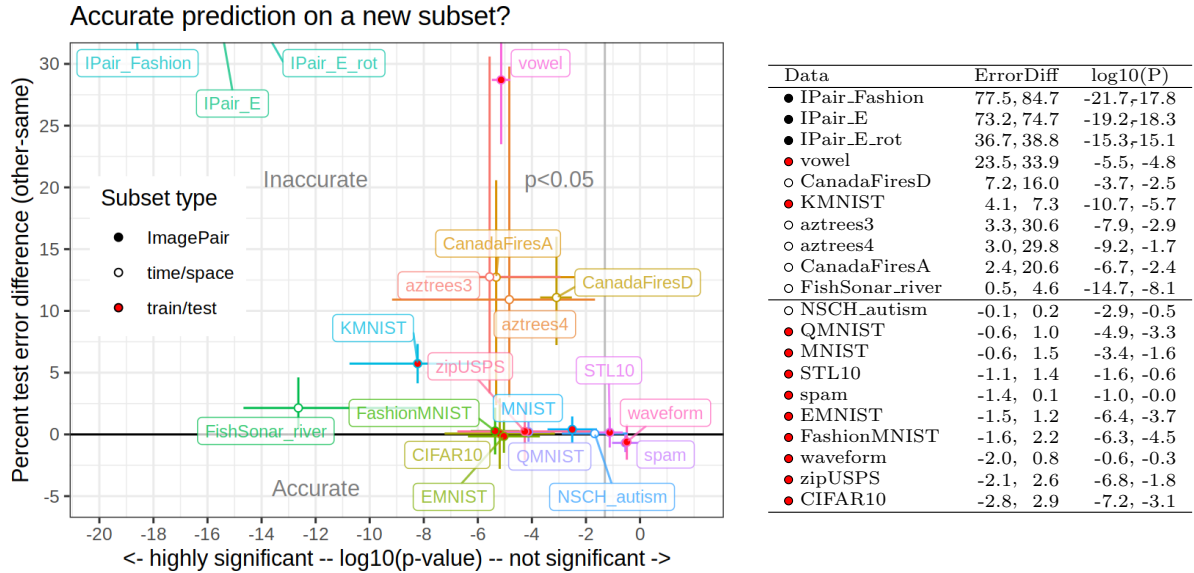
\begin{figure}[t]
    \centering
    \parbox{0.61\textwidth}{
    \includegraphics[width=0.59\textwidth]{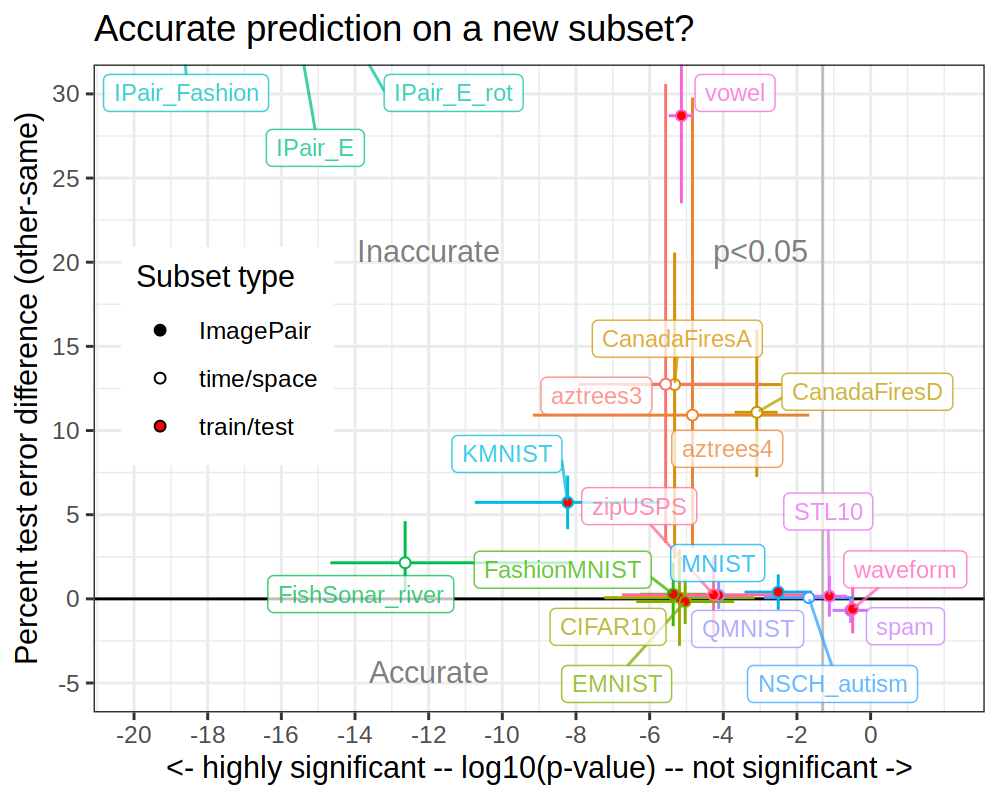}
    }
    \parbox{0.38\textwidth}{
    \scriptsize
\begin{tabular}{lll}
  \hline
Data & ErrorDiff & log10(P) \\ 
  \hline
\tikz\draw[black,fill=black] (0,0) circle (.5ex); IPair\_Fashion & $\parbox{0.5cm}{\rightline{77.5}},\parbox{0.5cm}{\rightline{84.7}}$ & $\parbox{0.5cm}{\rightline{-21.7}},\parbox{0.5cm}{\rightline{-17.8}}$ \\ 
  \tikz\draw[black,fill=black] (0,0) circle (.5ex); IPair\_E & $\parbox{0.5cm}{\rightline{73.2}},\parbox{0.5cm}{\rightline{74.7}}$ & $\parbox{0.5cm}{\rightline{-19.2}},\parbox{0.5cm}{\rightline{-18.3}}$ \\ 
  \tikz\draw[black,fill=black] (0,0) circle (.5ex); IPair\_E\_rot & $\parbox{0.5cm}{\rightline{36.7}},\parbox{0.5cm}{\rightline{38.8}}$ & $\parbox{0.5cm}{\rightline{-15.3}},\parbox{0.5cm}{\rightline{-15.1}}$ \\ 
  \tikz\draw[black,fill=red] (0,0) circle (.5ex); vowel & $\parbox{0.5cm}{\rightline{23.5}},\parbox{0.5cm}{\rightline{33.9}}$ & $\parbox{0.5cm}{\rightline{-5.5}},\parbox{0.5cm}{\rightline{-4.8}}$ \\ 
  \tikz\draw[black,fill=white] (0,0) circle (.5ex); CanadaFiresD & $\parbox{0.5cm}{\rightline{7.2}},\parbox{0.5cm}{\rightline{16.0}}$ & $\parbox{0.5cm}{\rightline{-3.7}},\parbox{0.5cm}{\rightline{-2.5}}$ \\ 
  \tikz\draw[black,fill=red] (0,0) circle (.5ex); KMNIST & $\parbox{0.5cm}{\rightline{4.1}},\parbox{0.5cm}{\rightline{7.3}}$ & $\parbox{0.5cm}{\rightline{-10.7}},\parbox{0.5cm}{\rightline{-5.7}}$ \\ 
  \tikz\draw[black,fill=white] (0,0) circle (.5ex); aztrees3 & $\parbox{0.5cm}{\rightline{3.3}},\parbox{0.5cm}{\rightline{30.6}}$ & $\parbox{0.5cm}{\rightline{-7.9}},\parbox{0.5cm}{\rightline{-2.9}}$ \\ 
  \tikz\draw[black,fill=white] (0,0) circle (.5ex); aztrees4 & $\parbox{0.5cm}{\rightline{3.0}},\parbox{0.5cm}{\rightline{29.8}}$ & $\parbox{0.5cm}{\rightline{-9.2}},\parbox{0.5cm}{\rightline{-1.7}}$ \\ 
  \tikz\draw[black,fill=white] (0,0) circle (.5ex); CanadaFiresA & $\parbox{0.5cm}{\rightline{2.4}},\parbox{0.5cm}{\rightline{20.6}}$ & $\parbox{0.5cm}{\rightline{-6.7}},\parbox{0.5cm}{\rightline{-2.4}}$ \\ 
  \tikz\draw[black,fill=white] (0,0) circle (.5ex); FishSonar\_river & $\parbox{0.5cm}{\rightline{0.5}},\parbox{0.5cm}{\rightline{4.6}}$ & $\parbox{0.5cm}{\rightline{-14.7}},\parbox{0.5cm}{\rightline{-8.1}}$ \\ 
  \hline
  \tikz\draw[black,fill=white] (0,0) circle (.5ex); NSCH\_autism & $\parbox{0.5cm}{\rightline{-0.1}},\parbox{0.5cm}{\rightline{0.2}}$ & $\parbox{0.5cm}{\rightline{-2.9}},\parbox{0.5cm}{\rightline{-0.5}}$ \\ 
  \tikz\draw[black,fill=red] (0,0) circle (.5ex); QMNIST & $\parbox{0.5cm}{\rightline{-0.6}},\parbox{0.5cm}{\rightline{1.0}}$ & $\parbox{0.5cm}{\rightline{-4.9}},\parbox{0.5cm}{\rightline{-3.3}}$ \\ 
  \tikz\draw[black,fill=red] (0,0) circle (.5ex); MNIST & $\parbox{0.5cm}{\rightline{-0.6}},\parbox{0.5cm}{\rightline{1.5}}$ & $\parbox{0.5cm}{\rightline{-3.4}},\parbox{0.5cm}{\rightline{-1.6}}$ \\ 
  \tikz\draw[black,fill=red] (0,0) circle (.5ex); STL10 & $\parbox{0.5cm}{\rightline{-1.1}},\parbox{0.5cm}{\rightline{1.4}}$ & $\parbox{0.5cm}{\rightline{-1.6}},\parbox{0.5cm}{\rightline{-0.6}}$ \\ 
  \tikz\draw[black,fill=red] (0,0) circle (.5ex); spam & $\parbox{0.5cm}{\rightline{-1.4}},\parbox{0.5cm}{\rightline{0.1}}$ & $\parbox{0.5cm}{\rightline{-1.0}},\parbox{0.5cm}{\rightline{-0.0}}$ \\ 
  \tikz\draw[black,fill=red] (0,0) circle (.5ex); EMNIST & $\parbox{0.5cm}{\rightline{-1.5}},\parbox{0.5cm}{\rightline{1.2}}$ & $\parbox{0.5cm}{\rightline{-6.4}},\parbox{0.5cm}{\rightline{-3.7}}$ \\ 
  \tikz\draw[black,fill=red] (0,0) circle (.5ex); FashionMNIST & $\parbox{0.5cm}{\rightline{-1.6}},\parbox{0.5cm}{\rightline{2.2}}$ & $\parbox{0.5cm}{\rightline{-6.3}},\parbox{0.5cm}{\rightline{-4.5}}$ \\ 
  \tikz\draw[black,fill=red] (0,0) circle (.5ex); waveform & $\parbox{0.5cm}{\rightline{-2.0}},\parbox{0.5cm}{\rightline{0.8}}$ & $\parbox{0.5cm}{\rightline{-0.6}},\parbox{0.5cm}{\rightline{-0.3}}$ \\ 
  \tikz\draw[black,fill=red] (0,0) circle (.5ex); zipUSPS & $\parbox{0.5cm}{\rightline{-2.1}},\parbox{0.5cm}{\rightline{2.6}}$ & $\parbox{0.5cm}{\rightline{-6.8}},\parbox{0.5cm}{\rightline{-1.8}}$ \\ 
  \tikz\draw[black,fill=red] (0,0) circle (.5ex); CIFAR10 & $\parbox{0.5cm}{\rightline{-2.8}},\parbox{0.5cm}{\rightline{2.9}}$ & $\parbox{0.5cm}{\rightline{-7.2}},\parbox{0.5cm}{\rightline{-3.1}}$ \\ 
   \hline
\end{tabular}
}
    \caption{SOAK was used to compute mean test error differences (Other-Same) and p-values for each test subset, over 10 cross-validation folds.
    Line segments and table show min/max values over 2--4 test subsets in each data set; dot shows mean.
    Horizontal black line separates data sets by the degree of differences in learnable/predictable patterns: top 10 for large differences (min/max ErrorDiff both positive: inaccurate prediction on all new subsets) and bottom 10 for small differences (min ErrorDiff negative, max positive: accurate prediction on at least one new subset).
    }
    \label{fig:map_other}
\end{figure}

Our goal in analyzing these 3 data sets was to quantify our intuition that MNIST/EMNIST are more similar than MNIST/FashionMNIST.
We expected that the worst prediction accuracy should be in IPair\_Fashion, because the subsets are most different (images of digits/clothing), and that is what we observed (Figure~\ref{fig:map_other}, Other-Same test error differences of 77.5--84.7\%, $p<10^{-17}$). 
We expected an intermediate prediction accuracy for IPair\_E, because both subsets are images of digits (although not in the same orientation), and that is what we observed (Figure~\ref{fig:map_other}, Other-Same test error differences of 73.2--74.7\%, $p<10^{-18}$).
We expected best prediction accuracy for IPair\_E\_rot, because both subsets are images of digits (in the same orientation), and that is what we observed (Figure~\ref{fig:map_other}, Other-Same test error differences of 36.7--38.8\%, $p<10^{-15}$).
Overall, our SOAK analysis indicates that after training on MNIST, the predictions on EMNIST are better than on FashionMNIST, but still not as good as prediction on MNIST (positive Other-Same differences in Figure~\ref{fig:map_other}), which is consistent with the different method used to construct the EMNIST data \citep{EMNIST}.

\subsection{Analysis of data with predefined train/test subsets}

In analyzing these 11 data sets, our goal was to verify that most data sets have similar predefined train/test splits.
In examining Figures~\ref{fig:map_all}--\ref{fig:map_other}, we observed that most data sets with predefined train/test subsets were in the similar category (MNIST, FashionMNIST, QMNIST, STL10, CIFAR10, EMNIST, waveform, spam, zipUSPS), and two had clearly different subsets (vowel and KMNIST).
Our SOAK analysis indicates that the predefined train/test splits in vowel/KMNIST represent a problem which is not i.i.d., so is difficult for standard supervised learning algorithms.

\subsection{Conclusions and future work}
We presented Same/Other/All K-fold cross-validation (SOAK), which can be used to estimate similarity/differences between learnable/predictable patterns in data subsets.
We showed how SOAK can be used to gain insights about 20 real-world and benchmark data sets. 
We quantitatively verified that a model trained on MNIST digits is not as accurate on FashionMNIST (clothing), as on EMNIST (digits).
We were also able to verify that most benchmark data had similar predefined train/test subsets (except vowel/KMNIST).
We observed significant differences between learnable/predictable patterns in space/time subsets of real data sets. 
However, results for NSCH\_autism indicate it would be slightly beneficial to train a model on the combined data from different years (Figure~\ref{fig:map_all}).
Future work includes algorithms for efficiently searching the space of subsets to use for training, which is a combinatorial problem, but may be tractable with submodular optimization \citep{bach2013learning}.

\textbf{Broader Impacts:}
It is important to be able to assess the accuracy of machine learning algorithms in real-world situations, and our proposed SOAK algorithm is ideal for that purpose. 
Negative impacts include large computation times and energy usage required to train and compare models (similar to standard K-fold CV).

\textbf{Limitations:}
SOAK is only applicable to data sets with subsets of interest (this may not be true of all data sets), for which there is enough time to use K-fold cross-validation (we used a linear model, which is fast enough, but this may not be appropriate for other learning algorithms, especially deep neural networks with lots of parameters).

\textbf{Software, Documentation, and Reproducibility:}
We provide an R package, published on the CRAN, which implements SOAK in the mlr3 framework: \url{https://github.com/tdhock/mlr3resampling}.
Detailed documentation for the software, including example code, can be found in the vignettes published on CRAN: \url{https://cloud.r-project.org/web/packages/mlr3resampling/}.
We also provide a GitHub repository with code used to make the figures in this paper: \url{https://github.com/tdhock/cv-same-other-paper}.

\bibliographystyle{abbrvnat}
\bibliography{refs}

\end{document}